\documentclass{article}

\PassOptionsToPackage{numbers, compress}{natbib}
\usepackage[final]{nips_2018}

\usepackage[utf8]{inputenc} % allow utf-8 input
\usepackage[T1]{fontenc}    % use 8-bit T1 fonts
\usepackage{hyperref}       % hyperlinks
\usepackage{url}            % simple URL typesetting
\usepackage{booktabs}       % professional-quality tables
\usepackage{amsfonts}       % blackboard math symbols
\usepackage{nicefrac}       % compact symbols for 1/2, etc.
\usepackage{microtype}      % microtypography

\usepackage{graphicx}

\title{Using Multitask Learning to Improve 12-Lead Electrocardiogram Classification}

\author{
  J. Weston Hughes\thanks{corresponding authors: jwhughes@berkeley.edu; geoff.tison@ucsf.edu} \\
  EECS\\
  UC Berkeley\\
  \And
  Taylor Sittler MD \\
  Laboratory Medicine\\
  UCSF \\
  \And
  Anthony D. Joseph \\
  EECS\\
  UC Berkeley\\
  \AND
  Jeffrey E. Olgin MD \\
  Cardiology\\
  UCSF \\
  \And
  Joseph E. Gonzalez \\
  EECS\\
  UC Berkeley\\
  \And
  Geoffrey H. Tison MD$^*$ \\
  Cardiology\\
  UCSF
}

\begin{document}

\maketitle

\begin{abstract}
We develop a multi-task convolutional neural network (CNN) to classify multiple diagnoses from 12-lead electrocardiograms (ECGs) using a dataset comprised of over 40,000 ECGs, with labels derived from cardiologist clinical interpretations. Since many clinically important classes can occur in low frequencies, approaches are needed to improve  performance on rare classes. We compare the performance of several single-class classifiers on rare classes to a multi-headed classifier across all available classes. We demonstrate that the addition of common classes can significantly improve CNN performance on rarer classes when compared to a model trained on the rarer class in isolation. Using this method, we develop a model with high performance as measured by F1 score on multiple clinically relevant classes compared against the gold-standard cardiologist interpretation.
\end{abstract}

\section{Background}
\subsection{Electrocardiograms}
Electrocardiograms (ECGs) are recordings of the electrical activity of the heart measured by placing one or more electrodes (leads) on the skin of a patient at various locations. ECGs are used to non-invasively diagnose various cardiac abnormalities \cite{heden}. The output of an ECG is a sequence of voltages at each lead sampled hundreds of times per second which, when plotted in a standardized format, is usually interpreted by a cardiologist or other expert to diagnose various cardiac abnormalities ranging from rhythm disturbances to abnormal cardiac blood flow, which can signify a heart attack. The ECG is the most common diagnostic cardiac test, with hundreds of millions of ECGs recorded each year. Though single or several-lead wearable ECGs have become increasingly common allowing for continuous monitoring of fewer channels at a lower resolution \cite{ng}, the 12-lead ECG is commonly considered to be the non-invasive gold standard ECG test.
 
Given the critical importance of ECGs to clinical medicine and the large number of ECGs recorded worldwide, high accuracy in automatic classification systems is critical and has substantial clinical impact. For example, ER doctors in high-volume, high-stress situations may have increased error rates, which could be reduced through more accurate automated classification \cite{errors}. Existing non-machine learning methods use wavelet, Fourier or other heuristic methods on select subsets of leads to classify specific abnormalities, but in practice can have low accuracy, particularly on rhythm diagnoses \cite{docerrors, docerrors2}. Recently, there has been success in applying convolutional neural networks (CNN) to the problem of classifying common arrhythmias in single-lead ECGs  using a large dataset \cite{ng}, but to our knowledge, there have been no demonstrations of using a CNN to detect a wide array of diagnoses on a large dataset of 12-lead ECGs. A critical barrier to training CNNs is that in most clinical datasets, many important classes are rare and can occur in very low frequencies. Approaches are needed to improve performance on rarer classes, ideally by leveraging existing data and labels.

\subsection{Rare, Difficult, and Hierarchical Classes}
The set of labels assigned to ECGs is heterogeneous and any number of potential diagnostic classes can be present in a given record. Some clinically significant diagnostic classes, such as complete heart block, appear in most clinical datasets (including ours) at very low frequencies. The optimal approach to train a model to identify a low frequency class in a large dataset remains an open question. One set of approaches focuses on finding lower-dimensional representations of the important details of the data, and using these representations for classification \cite{kingma}. Another approach to achieve quality representations, which we employ in this work, is through multi-task learning, where multiple heads of the same model are used for different classification tasks \cite{multisurvey}. This allows information to be shared throughout the model, through gradients originating from the loss of each prediction.

\begin{figure}
  \begin{center}
  \includegraphics[scale=.7, trim={0 .25cm 0 .25cm},clip]{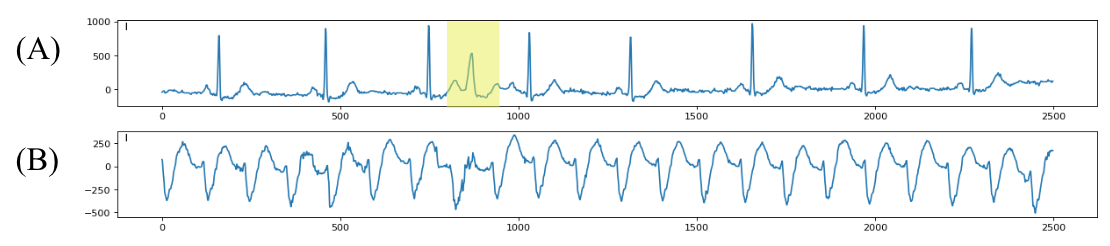}
  \end{center}
    \caption{Two examples in our training set (one of twelve channels is shown for each). (A) shows a premature ventricular complex during a single beat (highlighted), as well as bradycardia (characterized by a slow rate of beats); this record morphologically appears relatively normal. (B) shows ventricular tachycardia, characterized by a high rate of abnormal beats of a characteristic morphology.}
\end{figure}

As shown in Figure 1, ECG classes can vary in a variety of distinct domains, such as exhibiting normal morphology at high periodicity (e.g. supraventricular tachycardia), exhibiting abnormal morphology in a single beat (e.g. premature ventricular complex), or by varying in a combination of these domains (e.g. ventricular bigeminy). This means that a multi-task model must both take into account recurrent patterns across the entire input, as well as local events which appear at a single time point and which may only be present on a subset of the channels. Classes can also be hierarchical or correlated. For example, supraventricular, ventricular, and atrial tachycardia are all subsets of tachycardia (elevated heart rate) which originate from different parts of the heart. Many premature ventricular complexes occur along with sinus rhythm, while tachycardia and bradycardia (a low heart rate) tend not to co-occur. 

\section{Dataset}
This dataset consists of deidentified ECGs obtained during the course of clinical care at the University of California, San Francisco (UCSF) between 2010 and 2017. IRB approval was obtained for this study. ECGs are recorded in a standard 12-lead format (MUSE Version 9.0 SP4, GE Healthcare, Wauwatosa, WI). As part of routine clinical care, each ECG underwent initial analysis by the GE MUSE software and each interpretation was subsequently changed or confirmed by a UCSF board-certified cardiologist. 

We downsampled all examples to 250Hz at the standard 12-lead ECG recording length of 10 seconds, resulting in a 12 channel by 2500 measurement input. Binary labels for 24 classes were extracted from cardiologist-confirmed text diagnoses fields through simple string matching which were hand-verified and spot checked. We re-sampled the dataset to improve class balance by preferentially adding examples from minority classes until there were 4000 examples of that class in the resampled dataset, or the class was depleted in the original dataset. This still resulted in over-representation of the common classes that overlapped with rarer classes. The resulting dataset consisted of 41,522 examples. This is a sufficient number to train a quality model on common classes, since many common classes exhibit high degrees of internal repetition.

\section{Methods}
We train a 34-layer convolutional neural net with residual connections \cite{resnet} on the data, closely following the architecture described in \citet{ng}. We input a 12 channel by 2500 measurement sequence of voltages into the model, and all convolutions are along the time axis. Every 4 convolutional layers we max pool with stride and pooling size 2, and every 8 layers we increase the number of channels by a factor of 2, starting from 64. Thus the output of our final layer is a tensor of shape $256\times 10$, which we flatten and use as the input to multiple independent logistic regressions on each classification task. Whereas \cite{ng} predicts a single label for each second of their input, we make multiple binary predictions over the entire input, optimizing a sum of sigmoid cross-entropy losses. We employ ReLU non-linearities \cite{relu}, batch normalization \cite{batch}, and L2 regularization on all layers. We trained all models for 96 epochs using ADAM \cite{adam} with default settings and fixed random seed.

In the setting of high class imbalance, accuracy and ROC scores can easily be fooled because of Simpson's paradox \cite{auc}. Even if a model performs well in classifying positives and classifying negatives, the number of false positives might still outweigh the number of true positives, reducing the precision of the classifier. In cases where false positives can lead to costly and invasive interventions, as is often the case in medicine, this is unacceptable. To examine model performance, we instead use the F1 score which equally weighs the precision and recall of a classifier \cite{f1book}. 

All versions of the model differ only in the last layer. We trained several single-class models which flatten the final convolutional output and perform logistic regression to give a score between 0 and 1 for each example. The multi-headed models work the same way, but multiple independent logistic regressions are applied to the final layer, one for each of the classes being predicted. 

\begin{figure}
  \begin{center}
  \includegraphics[scale=.5,trim={0 .75cm 0 .25cm},clip]{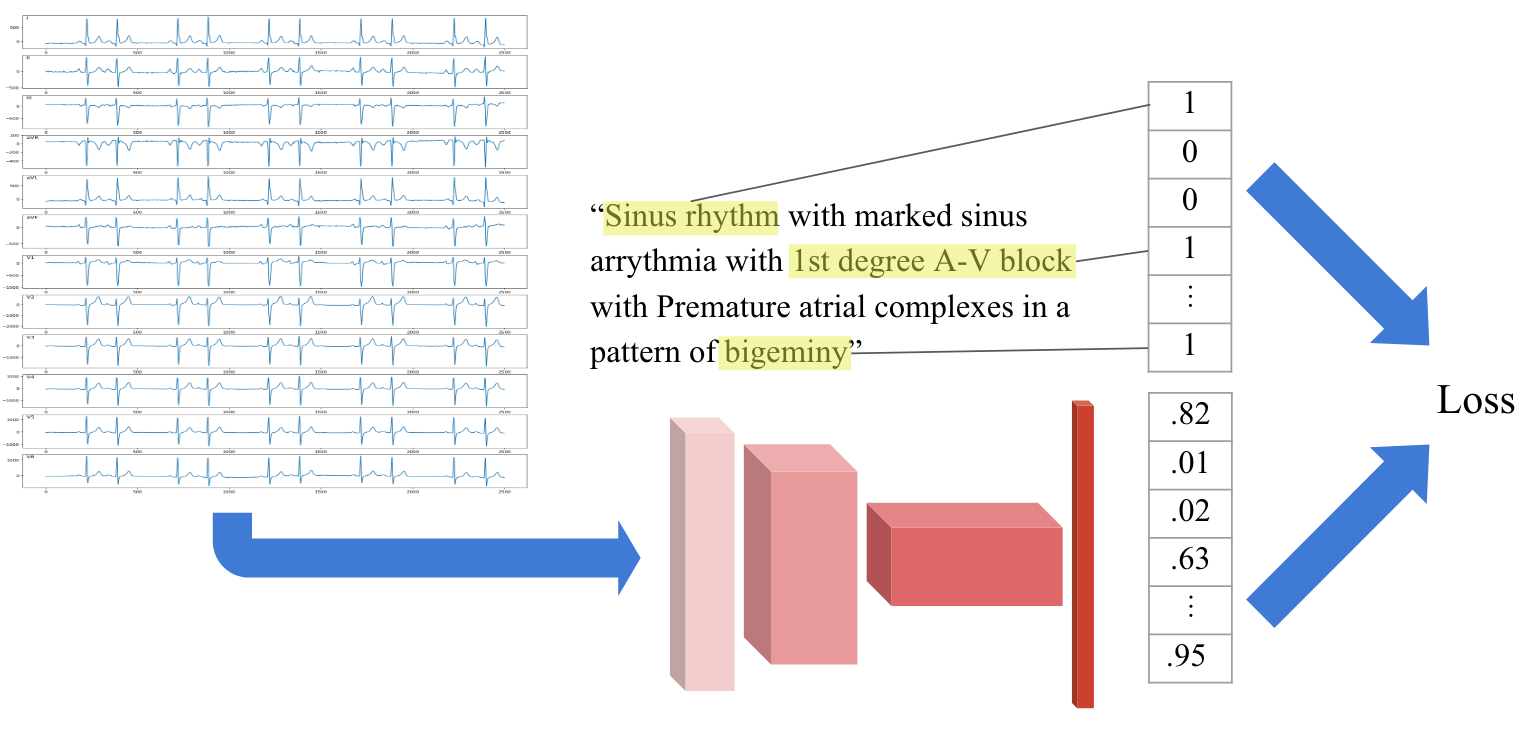}
  \end{center}
    \caption{Our convolutional model takes 2500 voltage measurements from 12 ECG channels and outputs multiple binary predictions on a set of non-mutually exclusive classes.}
\end{figure}

\section{Results}

\begin{table}[!tp]
  \caption{Results for multiple identical models trained on increasing numbers of classification tasks. Single-headed models were trained for five classes. One 2-class model (A) was trained on 2nd Degree AV block (Mobitz I) and 1st Degree AV block, and a multi-class model (B) was trained on all available classes. The multi-headed models in all cases perform as well or better than single-headed models; the improvement is more pronounced for rarer classes. In the case of 2nd Degree AV Block (Mobitz I), the addition of just one common class provides as much improvement as the addition of all classes. The best F1 Score for each class is bolded.}
  \label{sample-table}
  \centering
  \makebox[\textwidth][c]{
  \begin{tabular}{lcccccccc}
    \toprule
    F1 Scores\\
    \midrule
    Class  & Count &\multicolumn{5}{c}{Single-headed Models} & \multicolumn{2}{c}{Multi-headed Models}    \\
    ~ & ~ & ~ & ~ & ~ & ~ & ~ & A & B\\
    \midrule
    {\bf Rare Classes}\\
    Complete Heart Block & 204 & .09 & - & - & - & - & - & {\bf .62}\\
    Atrioventricular Nodal Reentry Tachycardia & 372 & - & .39 & - & - & - & - & {\bf .49}\\
    2nd Degree AV Block (Mobitz I) & 626 & - & - & .58 & - & - & {\bf .73} & {\bf .73}\\
    \midrule
    {\bf Common Classes}\\
    Atrial Fibrillation & 4719 & - & - & - & .84 & - & - & {\bf .85}\\
    Ectopic Atrial Rhythm & 1710 & - & - & - & - & {\bf .74} & - & {\bf .74}\\
    1st Degree AV Block & 4673 & - & - & - & - & - & {\bf .83} & {\bf .83}\\
    Sinus Rhythm & 25288 & - & - & - & - & - & - & {\bf .91}\\
    Atrial Flutter & 4072 & - & - & - & - & - & - & {\bf .80}\\
    Premature Atrial Complexes & 4905 & - & - & - & - & - & - & {\bf .78}\\
    Premature Ventricular Complexes & 4905 & - & - & - & - & - & - & {\bf .84}\\
    Fusion Complex & 2683 & - & - & - & - & - & - & {\bf .45}\\
    Bigeminy & 1445 & - & - & - & - & - & - & {\bf .87}\\
    
    \bottomrule
  \end{tabular}
  }
\end{table}

 We trained five single-class and two multi-class models, all using the same data, architecture, and training approach except for the set of labels being predicted (Table 1). We trained single-class models for three rare classes (Complete Heart Block, Atrioventricular Nodal Reentry Tachycardia, and 2nd Degree AV Block Mobitz I) and two common classes (Atrial Fibrillation and Ectopic Atrial Rhythm). We found that model performance by F1 score on rare classes was significantly improved by the addition of other classes. 

The primary finding of this work is that multi-headed classifiers generally outperform single-headed classifiers for any given class, with more pronounced improvements for rarer classes. With only a few hundred examples out of >40,000 records, rare classes have lower F1 scores by single-headed classifiers, whereas the multi-headed classifier performs significantly better on these rare classes. For more common classes, the single-headed model performs relatively well and the improvement provided by the multi-headed model is lower. We also found that in the case of the rare class of 2nd Degree AV Block (Mobitz I), the addition of a single related and common class (1st Degree AV Block) provides the same gain in performance as the addition of many classes. This suggests that the threshold of necessary supervision to achieve a well-performing model can be relatively low.

\section{Discussion}
In this work we demonstrate that a multi-headed classifier for 12-lead ECG data trained on a combination of rare and common classes performs significantly better than single-headed classifiers individually trained on rare classes. We use this method to develop what we believe is the first reported high-performing multitask CNN for 12-lead ECG data for multiple medically important diagnoses. Our results suggest that the addition of more classes improves the quality of the representation learned by higher layers of the model by providing increased supervision and more examples to influence the choice of filters throughout the model. The success of the two-headed model (A), which improved performance on the rare class by including a related, common class, demonstrates that even relatively small increases in supervision can provide substantial increases in model performance. In this case, the improvement was equivalent to that gained by the addition of many more less-related classes.

Our approach has the potential to improve performance when applying CNN models to clinical datasets, such as 12-lead ECG data, where certain clinically important diagnostic classes may occur with low frequency. We show that the addition of even one related classification task can improve performance for the rare class. While this approach was effective for classes appearing more than 200 times, it is worth noting that we have not observed similar performance gains for very rare classes, such as those with fewer than 100 examples. Future directions may include training a multi-headed model with many features, freezing it, and retraining the last layer using a small dataset of extremely rare classes and negatives. Other approaches include few-shot learning techniques to solve the problem of training a model with some common and some extremely rare classes \cite{Sontag}.

\small
\bibliographystyle{plainnat}
\bibliography{references}

\end{document}